\documentclass[letterpaper, 10 pt, conference]{ieeeconf}  

\IEEEoverridecommandlockouts                              

\usepackage{graphicx}
\usepackage{amsmath}
\usepackage{amssymb}
\usepackage{booktabs}

\usepackage{multirow}
\usepackage{multicol}
\usepackage{makecell}
\usepackage{subcaption}
\usepackage{bm}

\usepackage{diagbox}

\title{\LARGE \bf
Category-Adaptive Label Discovery and Noise Rejection \\ for Multi-label Image Recognition with Partial Positive Labels
}

\author{Tao Pu$^{1}$, Qianru Lao$^{1}$, Hefeng Wu$^{1}$, Tianshui Chen$^{2}$ and Liang Lin$^{1}$
\thanks{Tao Pu and Qianru Lao contribute equally to this work and share first authorship.}
\thanks{$^{1}$Tao Pu, Qianru Lao, Hefeng Wu, and Liang Lin are with the School of Computer Science and Engineering, Sun Yat-Sen University, Guangzhou, Guangdong, China
        (email: putao3@mail2.sysu.edu.cn; estherbear17@gmail.com; wuhefeng@mail.sysu.edu.cn; linliang@ieee.org).}
\thanks{$^{2}$Tianshui Chen is with the Guangdong University of Technology, Guangzhou, Guangdong, China
        (email: tianshuichen@gmail.com).}
}

\begin{document}

\maketitle
\thispagestyle{empty}
\pagestyle{empty}

\begin{abstract}
As a promising solution of reducing annotation cost, training multi-label models with partial positive labels (MLR-PPL), in which merely few positive labels are known while other are missing, attracts increasing attention. Due to the absence of any negative labels, previous works regard unknown labels as negative and adopt traditional MLR algorithms. To reject noisy labels, recent works regard large loss samples as noise but ignore the semantic correlation different multi-label images. In this work, we propose to explore semantic correlation among different images to facilitate the MLR-PPL task. Specifically, we design a unified framework, \textbf{Category-Adaptive Label Discovery and Noise Rejection}, that discovers unknown labels and rejects noisy labels for each category in an adaptive manner. The framework consists of two complementary modules: (1) Category-Adaptive Label Discovery module first measures the semantic similarity between positive samples and then complement unknown labels with high similarities; (2) Category-Adaptive Noise Rejection module first computes the sample weights based on semantic similarities from different samples and then discards noisy labels with low weights. Besides, we propose a novel category-adaptive threshold updating that adaptively adjusts the threshold, to avoid the time-consuming manual tuning process. Extensive experiments demonstrate that our proposed method consistently outperforms current leading algorithms.

\end{abstract}


\begin{figure}[!t] 
  \centering
  \includegraphics[width=0.95\linewidth]{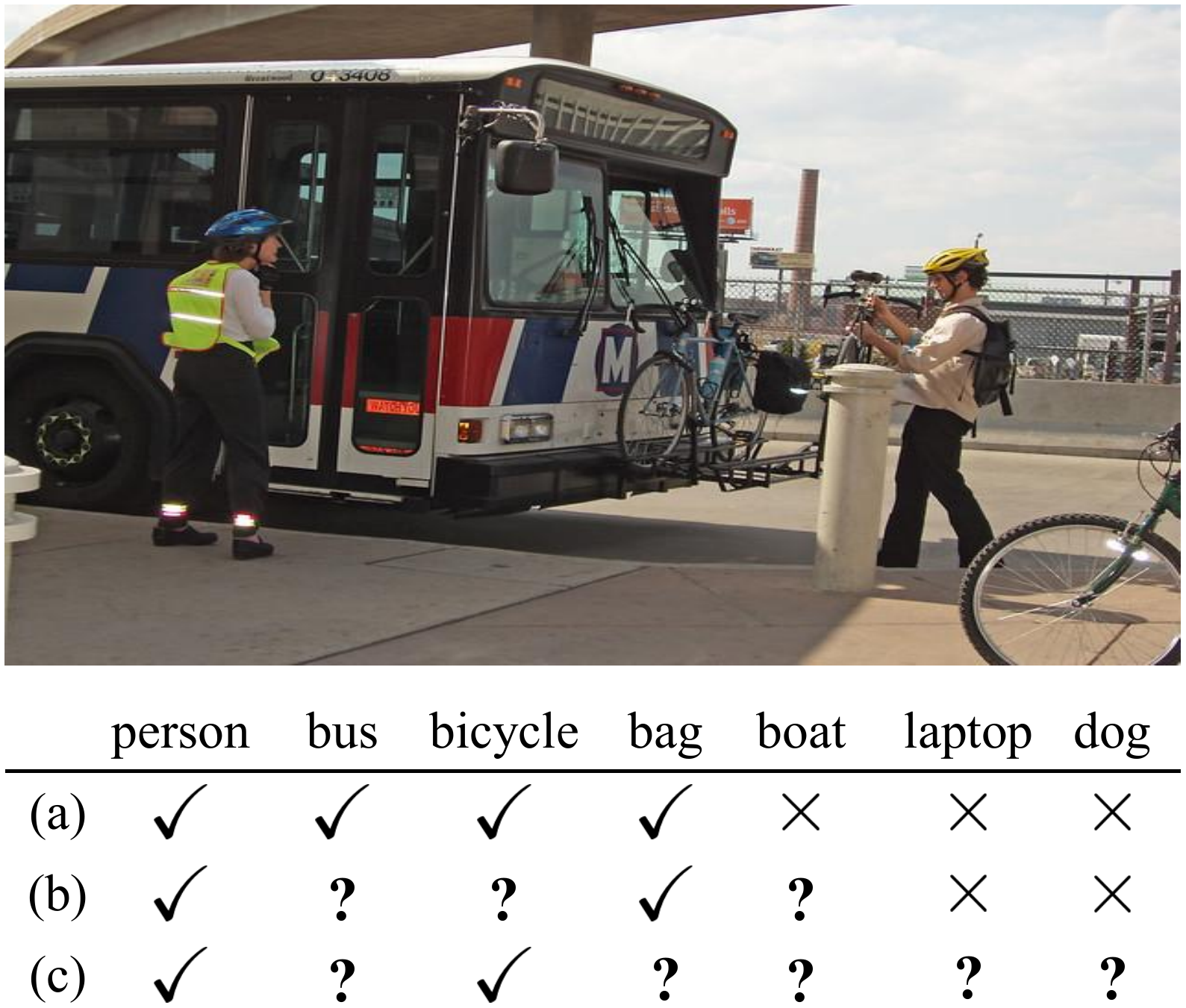}
  \caption{Example of the multi-label image with (a) full annotations, (b) partial annotations, and (c) partial positive annotations, in which $\checkmark$ represents the corresponding category exists, $\times$ represents it does not exist, and $?$ represents it is unknown. In partial positive multi-label learning (PPML), for this image, only part positive labels (\textit{person}, \textit{bus}) are annotated.}
  \label{fig:task}     
\end{figure}

\section{Introduction}
In last decades, lots of efforts \cite{Chen2019SSGRL,Wu2020AdaHGNN,Chen2021P-GCN,Chen2022KGGR} are dedicated to the task of multi-label image recognition (MLR), as it benefits various applications ranging from object retrieval \cite{Huang2021RA-L} and scene recognition \cite{Wald2018RA-L} to human activity recognition \cite{Mojarad2020RA-L} and pedestrian detection \cite{Kim2021RA-L}. Despite achieving impressive progress, existing algorithms extremely depend on collecting large-scale clean and complete multi-label datasets. That is not only extremely time-consuming but also is extremely expensive, especially when the number of categories and images increase in the dataset. To address this problem, weakly supervised multi-label image recognition (WSMLR) \cite{Bucak2011Incomplete,Sun2010Weak,Yang2016Missing} has been taken into consideration. Among different settings in WSMLR, the setting of merely few positive labels are annotated while other labels are missing per image (i.e., MLR-PPL) attracts increasing attention in recent years. That mainly because there exists of millions of images with tags provided by users in the internet and utilizing these web data to train multi-label models can obviously reduce the annotation cost and human intervention. As shown in Figure \ref{fig:task}, the MLR-PPL task not only faces the problem of losing lots of supervision signals, but also faces the dilemma lack of any negative labels that results model to always predict positive labels.

As aforementioned, due to the absence of any negative training examples in the partial positive labels setting, directly training multi-label recognition model always collapses to the trivial ``always predict positive” solution. Previous works \cite{Bucak2011Incomplete, Sun2010Weak, Yang2016Missing} propose a naive training strategy, Assume Negative (AN), where all unannotated labels are regarded as negative. Although such a training strategy avoid the this dilemma, these methods could suffer from severe performance drop because lots of positive labels be wrongly annotated as negative. More recent work \cite{Cole2021ROLE,Kim2022LargeLoss} propose to reject large loss samples to prevent model from memorizing the noisy labels. Despite achieving impressive progress, directly regrading the large loss samples as noisy labels easily drop lots of hard positive samples. Thus this method suffer from obvious performance drop, especially when the known positive label proportion is few.

Similar with the single counterpart, the multi-label images contains strong semantic correlation cross different images, as verified in recent work \cite{Chen2022SST}. Specifically, the objects of the same category in different images share similar visual appearance, thus we can infer the unannotated labels based on its semantic similarity among different images. However, the multi-label images not only contain multiple objects of diverse semantic categories which have variant sizes and scatters over the whole image, but also contain complex and varied backgrounds. Thus, the image-specific semantic similarity hard to precisely present the cross-image seamntic correlation of each category. To address this problem, we design a unified framework, Category-Adaptive Labels Discovery and Noise Rejection, that consists of two complementary modules (i.e., CALD and CANR modules) to complement unknown labels and discards noisy labels. As verified in recent work \cite{Chen2022SST}, any two instance object, belonged same category, in different images always share similar visual appearance. 

To be specific, in the proposed framework, we first introduce a category-specific semantic learning (CSSL) module \cite{Chen2019SSGRL,Chen2022SST,Pu2022SARB} that guides model learning semantic representation of each category by incorporating category semantics. Then, we design a category-adaptive label discovery (CALD) module thatfirst measures the semantic similarity between positive samples and then adaptively complement unknown labels with high similarities. Meanwhile, we dsign a category-adaptive noise rejection (CANR) module that first computes the sample weights based on semantic similarities from different samples and then adaptively discards noisy labels with low weights. Besides, we propose a novel category-adaptive threshold updating that adaptively adjusts the threshold, thus the proposed framework can achieve the optimal performance in different datasets and different proportion settings.

The contributions of this work are summarized into three folds:
\begin{itemize}
  \item We propose a novel perspective to train multi-label model with partial positive labels by exploring category-specific cross-image semantic correlation. To achieve this, we design a unified framework, Category-Adaptive Labels Discovery and Noise Rejection, that consists of two complementary modules (i.e., CALD and CANR modules) to complement unknown labels and discards noisy labels.
  \item We propose a novel category-adaptive threshold updating that adaptively adjusts the threshold of each category that avoids extremely time-consuming and laborious manual tuning in different datasets and different known positive label proportions.
  \item We conduct extensive experiments on several large-scale MLR datasets, including Microsoft COCO \cite{Lin2014COCO}, Visual Genome \cite{Krishna2017VG} and Pascal VOC 2007 \cite{Everingham2010Pascal}, to demonstrate the effectiveness of our proposed framework. We also perform ablative studies to analyze the contribution of each module for better understanding.
\end{itemize}

\begin{figure*}[!t]
   \centering
   \includegraphics[width=0.95\linewidth]{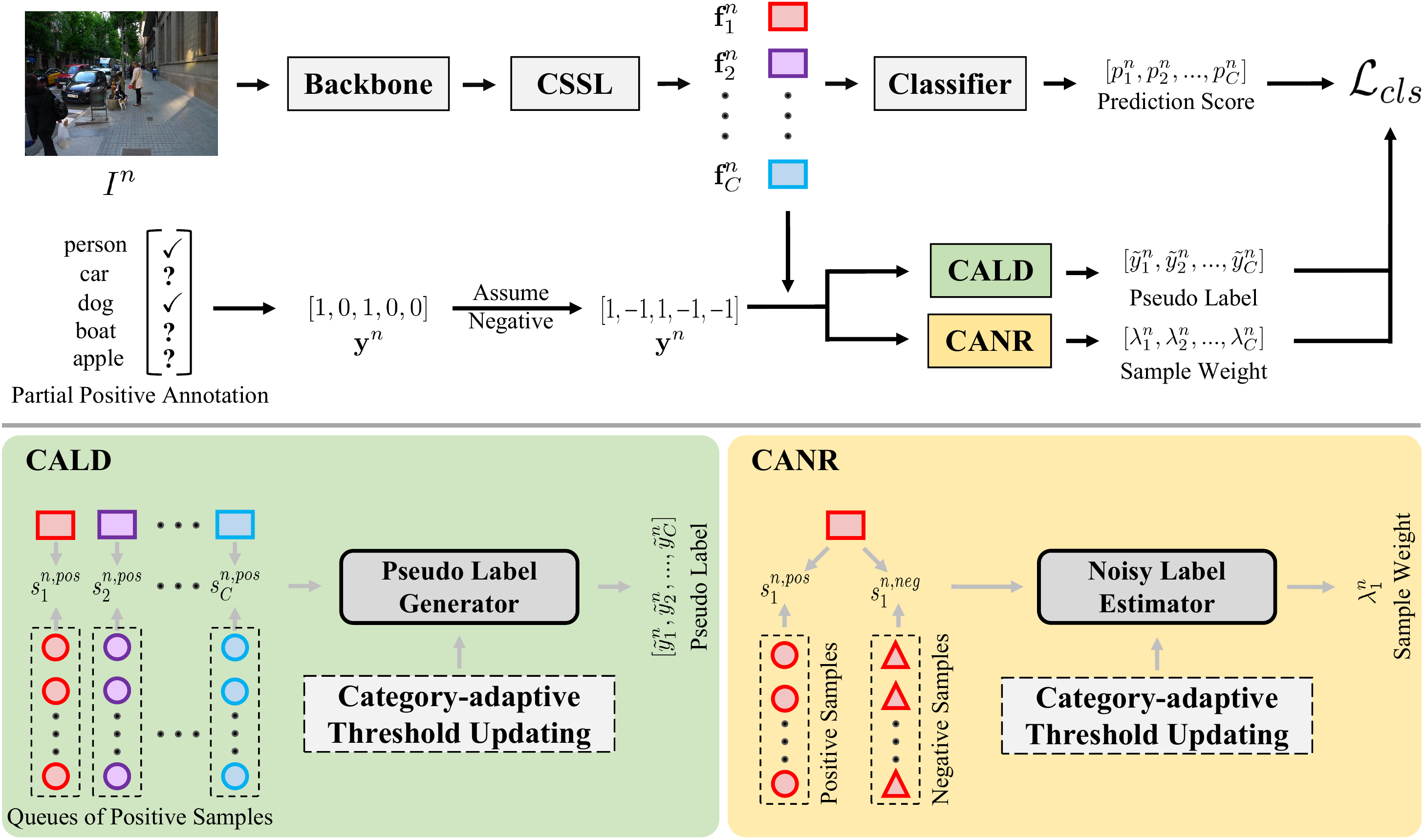}
   \caption{An overall illustration of the proposed category-adaptive label discovery and noise rejection framework. The upper part is the overall pipeline that consists of the CALD and CANR modules to generate pseudo labels and sample weights respectively. Due to the absence of any negative labels, we start our method with Assume Negative (AN) where all unknown labels are regarded as negative. The lower part is the detailed implementations of the CALD and CANR modules. The CALD module first measures the semantic similarities from the queue of positive samples and then complement unknown labels with high similarities. Meanwhile, the CANR module first computes the sample weights based on semantic similarities from different samples and then discards noisy labels with low weights.}
   \label{fig:framework}
\end{figure*}

\section{Related Work}
To solve this task, various works propose to capture discriminative local regions for feature enhancement by object proposal algorithms \cite{Wei2016HCP,Yang2016Exploit} or visual attention mechanisms \cite{Ba2014Multiple,Chen2018Recurrent}. Apart from that, another line of works propose to capture label dependencies to regularize training multi-label recognition models \cite{Chen2019ML-GCN,Chen2019SSGRL,Wu2020AdaHGNN,Ye2020ADD-GCN,Pu2022SRDL}. These works introduce structured graph to explicitly model the label dependencies to to facilitate the multi-label image recognition. To be specific, Chen et al. \cite{Chen2019ML-GCN} propose to build a directed graph over the object labels to model label dependencies in multi-label image recognition. And Chen et al. \cite{Chen2019SSGRL} propose to decouple category-specific semantic representation and pass message among different categories. However, despite achieving impressive progress, these leading algorithms always depend on collecting large-scale clean and complete multi-label datasets (e.g., Visual Genome \cite{Krishna2017VG}, MS-COCO \cite{Lin2014COCO} and Pascal VOC \cite{Everingham2010Pascal}) to learn discriminative feature representation. That is not extremely time-consuming but also is extremely expensive, especially when the number of categories and images increase in the dataset.

To alleviate the annotation dilemma, some works propose to learn multi-label recognition models with partial positive labels, i.e., merely some positive labels are annotated. As a naive training way, some works \cite{Bucak2011Incomplete, Sun2010Weak, Yang2016Missing} propose to train model with Assume Negative (AN) where all unannotated labels are regarded as negative. Although such a training strategy avoid the degenerate ``always predict positive" solution, these methods could suffer from severe performance drop because lots of positive labels be wrongly annotated as negative. To overcome this issue, some works propose to correct the wrongly annotated labels by exploring label dependencies. Xu et al. \cite{Xu2013SpeedUp} propose a novel theory of matrix completion that explicitly explore the side information to reduce the requirement on the number of observed entries. Wu et al. \cite{Wu2015MixedGraph} propose a unified model of label dependencies by constructing a mixed graph. Apart from these methods, some other works propose to adjust the weight of each annotation or the correct the wrong labels base on prediction scores. \cite{Durand2019CVPR} train a model with partial labels and adopt a curriculum learning approach to label some unannotated easy samples using the model prediction. Huynh et al. \cite{Huynh2020CVPR} utilize the image similarity to predict missing labels. Despite achieving impressive progress, existing works still rely on lots of annotated positive labels for image or similarity model training, which results in poor performance when the known positive label proportions decrease to a low level.

Recently,  Kim et al. \cite{Kim2022LargeLoss} propose to regard large loss samples as noisy labels to prevent model from memorizing noisy labels. However, it easily drops lots of hard positive sample and thus results sub-optimal performance. Different from this method, Chen et al. \cite{Chen2022SST} propose to explore cross-image semantic correlation to complements unknown labels. Despite achieving great progress, the threshold of generating pseudo labels in this method need exhaustive search which is impossible for different datasets and different known positive label proportions. Different from these prior works, our proposed framework explores category-specific semantic correlation among different multi-label images to complements unknown labels and reject noisy labels for supervising the training of multi-label model. Besides, we propose a novel category-adaptive threshold updating that adaptively adjusts the optimal threshold, to avoid the time-consuming manual tuning process Thus, our proposed framework can obtains consistently well performance on all known positive label settings.


\section{Method}
\noindent{\textbf{Overview.}} In this section, we first introduce the preliminary of multi-label image recognition with partial positive labels and then describe in detail our proposed framework that explores category-specific cross-image semantic correlation to complement unknown labels and discard noisy labels. The proposed framework consists of two complementary modules that perform category-adaptive label discovering and category-adaptive noise rejecting respectively, i.e., the CALD and CANR modules. Specifically, the CALD module measures the semantic similarity among feature representation of the same category from queue of positive samples, and then utilizes these semantic similarity to generate pseudo labels. Meanwhile, the CANR module computes the sample weights based on semantic similarities from different samples, and then discards noisy labels with low weights.  Different from previous works \cite{Chen2022SST, Kim2022LargeLoss}, we design the category-adaptive threshold updating to adaptively adjust the thresholds of each category in CALD and CANR modules that avoids extremely time-consuming and laborious manual tuning. An overall illustration is presented in Figure \ref{fig:framework}.

\subsection{Preliminary}
In the multi-label image recognition setting, each image $I^n$ is associated with a vector of labels $\mathbf{y}^n = \{y^n_1, \cdots, y^n_C\}\in \{-1, 1\}^C$, where $y^n_c$ is assigned to 1 if label $c$ exists in the image $I^n$ and assigned to -1 if it does not exist. $N$ is the total number of training samples in the training set $\mathcal{D}=\{(I^1,\textbf{y}^1), ..., (I^N,\textbf{y}^N)\}$, $C$ is the total number of categories. The loss function of $n$-th sample in this setting as follows:
\begin{equation}
    \begin{aligned}
    \ell(\mathbf{y}^n, \mathbf{p}^n) &= -\sum^{C}_{c=1} (\mathbf{1}[y^n_c=1] \log(p^n_c) \\
    &+ \mathbf{1}[y^n_c=-1] \log(1-p^n_c)),
    \end{aligned}
\end{equation}
where $\mathbf{1}[\cdot]$ is an indicator function whose value is 1 if the argument is positive and is 0 otherwise, $p^n_c$ is corresponding prediction score for category $c$ in image $I^n$.

In partial positive multi-label learning, merely few positive labels are annotated per image while other labels are unannotated. Follow previous works \cite{Durand2019CVPR,Huynh2020CVPR,Chen2022SST,Pu2022SARB}, we change the value space of target $\mathbf{y}^n$ to $\{0, 1\}$, where value is assigned to 0 if corresponding label is unknown. In this partial positive label setting, directly utilize above loss function to supervise model is impractical because the absence of any negative labels makes model overfit on these positive labels and always predict positive. Thus we adopt ``assume negative" (AN) loss \cite{Cole2021ROLE}, i.e., assume unobserved labels are negative. The objective function can be defined as
\begin{equation}
    \begin{aligned}
    \ell^{AN}(\mathbf{y}^n, \mathbf{p}^n) &= -\sum^{C}_{c=1} (\mathbf{1}[y^n_c=1] \log(p^n_c) \\
    &+ \mathbf{1}[y^n_c=0] \log(1-p^n_c)).
    \end{aligned}
\end{equation}

\subsection{Category-Specific Semantic Learning} 
Given an input image $I^n$, we first employ ResNet-101 \cite{He2016ResNet} as the backbone network to extract global feature maps $\textbf{f}^n$, and then introduce a category-specific semantic learning (CSSL) module that incorporates category semantics to learn semantic representation of category $c$
\begin{equation}
    \mathbf{f}^n_c =  \phi_{cssl}(\textbf{f}^n, \mathbf{u}_c),
\end{equation}
where $\mathbf{u}_c$ is semantic embedding of category $c$. We repeat the above process and obtain the semantic representations of all category $[\textbf{f}^n_1, \textbf{f}^n_2, \cdots, \textbf{f}^n_C]$. There are different algorithms to implement the CSSL module, including semantic decoupling proposed in \cite{Chen2019SSGRL} and semantic attention mechanism proposed in \cite{Ye2020ADD-GCN}. Following current works \cite{Chen2022SST,Pu2022SARB}, we adopt the semantic decoupling to implement the CSSL module. 

After that, we use a linear classifier followed by a sigmoid function to compute the prediction score vectors $\textbf{p}^n = \{p^n_1, \cdots, p^n_C\}$. Besides, based on the learned category-specific semantic representation, the CALD and CANR modules are used to adaptively complement unknown labels and discard noisy labels. We introduce these two modules in the following.

\subsection{Category-Adaptive Label Discovery} 
As verified in recent work \cite{Chen2022SST}, any two instance object, belonged same category, in different images always share similar visual appearance. Therefore, previous work \cite{Huynh2020CVPR} propose to utilize the image similarity to predict unannotated labels from other images which share similar visual appearance. However, in the multi-label scenario, the image-specific semantic similarity hard to present the cross-image semantic correlation of each category. To alleviate this dilemma, we design the CALD module that first measures the semantic similarity between positive samples and then adaptively complement unknown labels with high similarities.

For each category $c$ of images $I^n$ and $I^m$, we utilize the cosine distance between these corresponding semantic representation to compute their semantic similarity, following previous work \cite{Chen2022SST}, formulated as
\begin{equation}
    s^{n,m}_c=cosine(\textbf{f}^n_c, \textbf{f}^m_c) = \frac{\textbf{f}^n_c\cdot\textbf{f}^m_c}{||\textbf{f}^n_c||\cdot||\textbf{f}^m_c||}.
\end{equation}
Suppose the positive label of category $c$ is missing, we propose to utilize the semantic similarity between this sample and another positive sample to complement this unknown label:
\begin{equation}
    \tilde{y}^{n}_{c}=\textbf{1}[s_{c}^{n,m} \cdot y^{m}_{c} \ge \theta^{pos}_c],
\end{equation}
where $\textbf{1}[\cdot]$ is an indicator function and $\theta^{pos}_{c}$ is a threshold to control the precision of complementing unknown labels of category $c$ and the detailed introduction about it in Section \ref{sec:CATU}. We repeat above process to correct potential positive labels for all unknown labels and combine it with original annotations, obtaining $\tilde{\textbf{y}}^{n}=\{\tilde{y}^{n}_1, \tilde{y}^{n}_2, \cdots, \tilde{y}^{n}_C\}$.

However, due to the scarcity of positive samples in the MLR-PPL task and the complex variations in data collecting process (e.g., pose, illumination, and etc.), merely utilizing a single semantic similarity to generate pseudo labels is extremely unstable. To address this problem, we construct a queue of positive samples $\mathcal{Q}^{pos}_c = \{m | y^{m}_{c}=1\}$, in which each image has positive label $c$. With this query, we compute the average positive semantic similarities $s^{n,pos}_c$ between $\textbf{f}^n_c$ and the corresponding positive semantic representation of the images in $\mathcal{Q}^{pos}_c$, and then generate pseudo label by
\begin{equation}
    s^{n,pos}_c = \frac{1}{|\mathcal{Q}^{pos}_c|}\sum_{\{m\in\mathcal{Q}^{pos}_c\}}s_{c}^{n,m},
\end{equation}
\begin{equation}
    \tilde{y}^{n}_{c}=\textbf{1}[s^{n,pos}_c \ge \theta^{pos}_{c}].
\end{equation}
where $\theta^{pos}_{c}$ is a threshold and the detailed introduction about it in Section \ref{sec:CATU}.

After that, we further utilize the pseudo label $\mathbf{\tilde{y}}^{n}_{c}$ to supervise the training of multi-label models
\begin{equation}
    \begin{aligned}
    \ell^{AN}(\mathbf{\tilde{y}}^{n}, \mathbf{p}^n) &= -\sum^{C}_{c=1} (\mathbf{1}[\tilde{y}^{n}_{c}=1] \log(p^n_c) \\
    &+ \mathbf{1}[\tilde{y}^{n}_{c}=0] \log(1-p^n_c)).
    \end{aligned}
\end{equation}

\subsection{Category-Adaptive Noise Rejection} 
Considering ``Assume Negative" (AN) loss inevitably leads to introducing much noisy labels, many works propose to reject these noisy labels based on prediction score or objective loss. Recently, Kim et al. \cite{Kim2022LargeLoss} propose to reject the large loss samples and gradually increasing the rejection rate during the training process. Despite achieving impressive progress, directly rejecting large loss samples easily results sub-optimal performance. To precisely reject the noisy labels, we design the CANR module that first computes the sample weights based on semantic similarities from different samples and then adaptively discards noisy labels with low weights

Suppose the label of category $c$ is missing and thus is annotated as negative labels by AN loss, we propose to utilize the semantic similarity between this sample and other samples to decide whether reject this label:
\begin{equation}
\lambda^n_c=
    \begin{cases}
         1, \quad y^n_c=1 \\
         \textbf{1}[(s^{n,pos}_c - \theta^{neg}_c)/\theta^{pos}_c - \theta^{neg}_c) > X], \quad y^n_c=0 \\
    \end{cases}
\end{equation}
where $\theta^{neg}_c$ is another threshold to control the precision of rejecting noisy labels of category $c$ and the detailed introduction about it in Section \ref{sec:CATU}. $X$ is a random variable that obeys a uniform distribution, i.e., $X \sim U[0, 1]$.

After that, we further introduce the sample weight term $\mathbf{\lambda}^n$ in the objective function
\begin{equation}
    \begin{aligned}
    \ell^{AN}(\mathbf{y}^n, \mathbf{p}^n, \mathbf{\lambda}^n) &= -\sum^{C}_{c=1} (\mathbf{1}[y^n_c=1] \log(p^n_c) \times \lambda^n_c \\
    &+ \mathbf{1}[y^n_c=0] \log(1-p^n_c) \times \lambda^n_c ).
    \end{aligned}
\end{equation}

\subsection{Category-Adaptive Threshold Updating} \label{sec:CATU}
As discussed above, $\theta^{pos}_c$ and $\theta^{neg}_c$ are two crucial hyperparameters that control the precision and recall of generating pseudo labels and rejecting nosiy label for category $c$. On the one hand, the optimal thresholds vary widely across classes. That because the training samples of different categories vary obviously and the difficult of learning compact representation of different categories vary. For instance, compared with \textit{person}, $\theta^{pos}_c$ of \textit{toothbrush} should be lower, because the positive samples of \textit{toothbrush} are fewer and the learned visual representation of \textit{toothbrush} is easily more dispersed. On the other hand, the optimal thresholds for different datasets and different known label proportions are obviously different. Thus, it is crucial to exhaustively search the optimal threshold for achieving the optimal performance. However, such a manual tuning process is extremely time-consuming and laborious.

Different from manual tuning thresholds \cite{Chen2022SST, Kim2022LargeLoss}, we propose a novel category-adaptive threshold updating (CATU) that adaptively adjusts the threshold of each category that avoids extremely time-consuming and laborious manual tuning in different datasets and different known positive label proportions. Specifically, we first count the average positive similarity $s^{pos}_c$ and average negative similarity $s^{neg}_c$ of each category $c$
\begin{gather}
    s^{pos}_c = \frac{1}{|D^{pos}_c|} \sum_{\{n \in D^{pos}_c\}} s^{n,pos}_c, \\
    s^{neg}_c = \frac{1}{|D^{neg}_c|} \sum_{\{n \in D^{neg}_c\}} s^{n,neg}_c, 
\end{gather}
where $D^{pos}_c$ is the subset of images in which the category $c$ is exist ,and $D^{neg}_c$ is the subset of images in which the category $c$ is not exist. After that, we utilize the counted similarity in previous epoch and in current epoch momentum average the similarity
\begin{gather}
    s^{pos}_c = \frac{b}{B} * s^{pos}_c + (1-\frac{b}{B}) * s^{pos'}_c,\\
    s^{neg}_c = \frac{b}{B} * s^{neg}_c + (1-\frac{b}{B}) * s^{neg'}_c,
\end{gather}
where $b$ is current batch index in current epoch, $B$ is total number of batches in each epoch, $s^{pos}_c$ and $s^{neg}_c$ are the average similarities in current epoch and $s^{pos'}_c$ and $s^{neg'}_c$ are the average similarities in previous epoch. Finally, we adaptively adjust the threshold $\theta^{pos}_c$ and $\theta^{neg}_c$, formulated as
\begin{gather}
    \theta^{pos}_c = max(s^{pos}_c, \quad \theta),\\
    \theta^{neg}_c = \frac{1}{2} (\theta^{pos}_c + s^{neg}_c), 
\end{gather}
where $\theta$ is a threshold.

\subsection{Optimization}
As previously discussed, we introduce respectively the classification loss with partial positive labels, with pseudo labels, with sample weights. Thus, the final classification loss can be defined as summing up these loss
\begin{equation}
\mathcal{L}_{cls}=\sum_{n=1}^N(\ell^{AN}(\mathbf{y}^n, \mathbf{p}^n)+\ell^{AN}(\mathbf{\tilde{y}}^n, \mathbf{p}^n) +\ell^{AN}(\mathbf{y}^n, \mathbf{p}^n, \mathbf{\lambda^n})).
\label{eq:total-classification-loss}
\end{equation}

As aforementioned, due to the scarcity of positive samples and the complex variations in data collecting process, the semantic representation of each category are not compact enough, especially when the proportion of known positive labels is few. To guide model learning more compact semantic representation, we propose to introduce a cross-image semantic learning loss \cite{Chen2022SST} for training, formulated as
\begin{equation}
    \mathcal{L}_{csl}=\sum_{n=1}^N\sum_{m=1}^N\sum_{c=1}^C \ell^{n,m}_c,
\end{equation}
where
\begin{equation}
    \ell^{n,m}_c=
        \begin{cases}
             1-s^{n,m}_c \quad & y_c^n=1,y_c^m=1\\
             1+s^{n,m}_c \quad & otherwise.
        \end{cases}
\end{equation}

Thus, the final loss can be defined as summing up the classification loss, the cross-image semantic learning loss
\begin{equation}
\mathcal{L}= \mathcal{L}_{cls} + \alpha \mathcal{L}_{csl}.
\label{eq:total-loss}
\end{equation}
Here, $\alpha$ is the balance parameters that ensure the two losses have comparable magnitude, so that we set $\alpha$ to 0.05 in the experiments.

\begin{table*}[!h]
  \centering
  \begin{tabular}{c|c|ccccccccc|c}
  \hline
  \centering Publication & Method & 10\% & 20\% & 30\% & 40\% & 50\% & 60\% & 70\% & 80\% & 90\% & Ave. mAP \\
  \hline
  \hline
  \centering \multirow{8}*{MS-COCO} & ResNet-101 & 11.0 & 19.6 & 27.9 & 37.7 & 46.0 & 51.1 & 54.9 & 57.3 & 61.1 & 40.7 \\
  \centering & SSGRL & 10.6 & 41.8 & 55.0 & 62.2 & 65.5 & 67.7 & 71.6 & 74.2 & 78.6 & 58.6 \\
  \centering & KGGR & 9.0 & 36.8 & 54.1 & 61.9 & 66.4 & 69.2 & 72.4 & 74.6 & 76.6 & 57.9 \\
  \centering & Focal Loss & 20.9 & 37.5 & 49.6 & 52.1 & 56.1 & 59.5 & 62.3 & 63.6 & 65.2 & 51.9 \\
  \centering & ASL & 19.4 & 43.2 & 50.2 & 59.7 & 69.8 & 68.1 & 66.5 & 64.2 & 61.2 & 55.8 \\
  \centering & CST & 16.5 & 35.5 & 57.4 & 64.0 & 69.9 & 74.1 & 75.4 & 77.6 & 78.6 & 61.0 \\
  \centering & LL-R & 21.6 & 38.9 & 53.8 & 61.1 & 65.2 & 67.5 & 69.8 & 70.6 & 71.6 & 57.8 \\
  \centering & ROLE & 19.6 & 39.4 & 51.8 & 58.7 & 62.9 & 66.0 & 68.0 & 69.8 & 71.2 & 56.4 \\
  \cline{2-12}
  \centering & Ours & \textbf{40.4} & \textbf{50.7} & \textbf{67.2} & \textbf{70.6} & \textbf{74.1} & \textbf{75.6} & \textbf{77.0} & \textbf{78.0} & \textbf{78.9} & \textbf{68.1} \\
  \hline
  \hline
  \centering \multirow{8}*{VG} & ResNet-101 & 10.4 & 17.2 & 21.0 & 23.6 & 26.2 & 28.5 & 30.0 & 31.5 & 32.7 & 24.6 \\
  \centering & SSGRL & 7.0 & 20.3 & 25.8 & 30.6 & 32.4 & 35.2 & 37.5 & 38.4 & 40.0 & 29.7 \\
  \centering & KGGR & 8.2 & 13.4 & 23.4 & 30.6 & 33.2 & 35.9 & 36.6 & 38.2 & 40.2 & 28.9 \\
  \centering & Focal Loss & 17.1 & 22.8 & 27.1 & 30.5 & 31.8 & 33.2 & 34.8 & 35.4 & 36.4 & 29.9 \\
  \centering & ASL & 16.4 & 24.8 & 29.8 & 33.6 & 36.3 & 37.2 & 38.4 & 39.0 & 39.2 & 32.5 \\
  \centering & CST & 12.9 & 22.5 & 27.3 & 31.1 & 34.7 & 37.1 & 38.1 & 42.1 & 43.4 & 32.1 \\
  \centering & LL-R & 16.8 & 22.9 & 29.2 & 32.2 & 34.7 & 36.1 & 37.2 & 38.5 & 39.2 & 31.9 \\
  \centering & ROLE & 16.9 & 24.0 & 29.0 & 32.3 & 34.2 & 35.5 & 36.9 & 38.0 & 38.9 & 31.7 \\
  \cline{2-12}
  \centering & Ours & \textbf{19.0} & \textbf{31.0} & \textbf{33.1} & \textbf{37.2} & \textbf{39.1} & \textbf{42.7} & \textbf{43.3} & \textbf{43.9} & \textbf{45.1} & \textbf{37.2} \\
  \hline
  \hline
  \centering \multirow{8}*{Pascal VOC 2007} & ResNet-101 & 20.4 & 35.9 & 52.3 & 74.9 & 80.6 & 84.6 & 86.4 & 89.0 & 90.6 & 68.3 \\
  \centering & SSGRL & 48.1 & 76.9 & 83.4 & 87.0 & 88.5 & 89.9 & 90.3 & \textbf{92.0} & 91.9 & 83.1 \\
  \centering & KGGR & 33.5 & 70.6 & 83.6 & 87.4 & 84.8 & 89.3 & 90.7 & 91.4 & 92.0 & 80.4 \\
  \centering & Focal Loss & 55.7 & 79.3 & 81.5 & 85.5 & 88.0 & 88.8 & 89.8 & 90.6 & 91.5 & 83.4 \\
  \centering & ASL & 71.7 & 80.4 & 85.5 & \textbf{87.6} & 88.9 & 89.9 & \textbf{91.0} & 91.5 & 92.0 & 86.5 \\
  \centering & CST & 70.2 & 80.6 & 85.4 & 87.6 & \textbf{89.0} & \textbf{90.2} & 90.9 & 91.4 & \textbf{92.6} & 86.4 \\
  \centering & LL-R & 19.2 & 36.3 & 62.0 & 76.6 & 81.1 & 84.5 & 86.2 & 89.2 & 89.3 & 69.4 \\
  \centering & ROLE & 21.8 & 53.8 & 73.3 & 82.1 & 84.8 & 87.0 & 88.8 & 90.5 & 91.1 & 74.8 \\
  \cline{2-12}
  \centering & Ours & \textbf{76.3} & \textbf{82.3} & \textbf{86.4} & 87.2 & 88.6 & 89.0 & 90.0 & 90.8 & 91.9 & \textbf{86.9} \\
  \hline
  \hline
  \end{tabular}
  \caption{Performance of our proposed framework and current state-of-the-art competitors for MLR-PPL on the MS-COCO, VG-200 and Pascal VOC 2007 datasets. The best results are highlighted in bold.}
  \label{tab:result}
\end{table*}

\begin{figure*}[!h] 
  \centering    
  \subcaptionbox*{}{
    \includegraphics[width=0.99\linewidth]{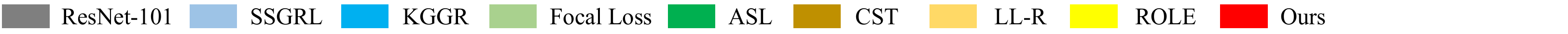}  
    }
  \vspace{-10pt}
  \subcaptionbox*{}{
    \includegraphics[width=0.3\linewidth]{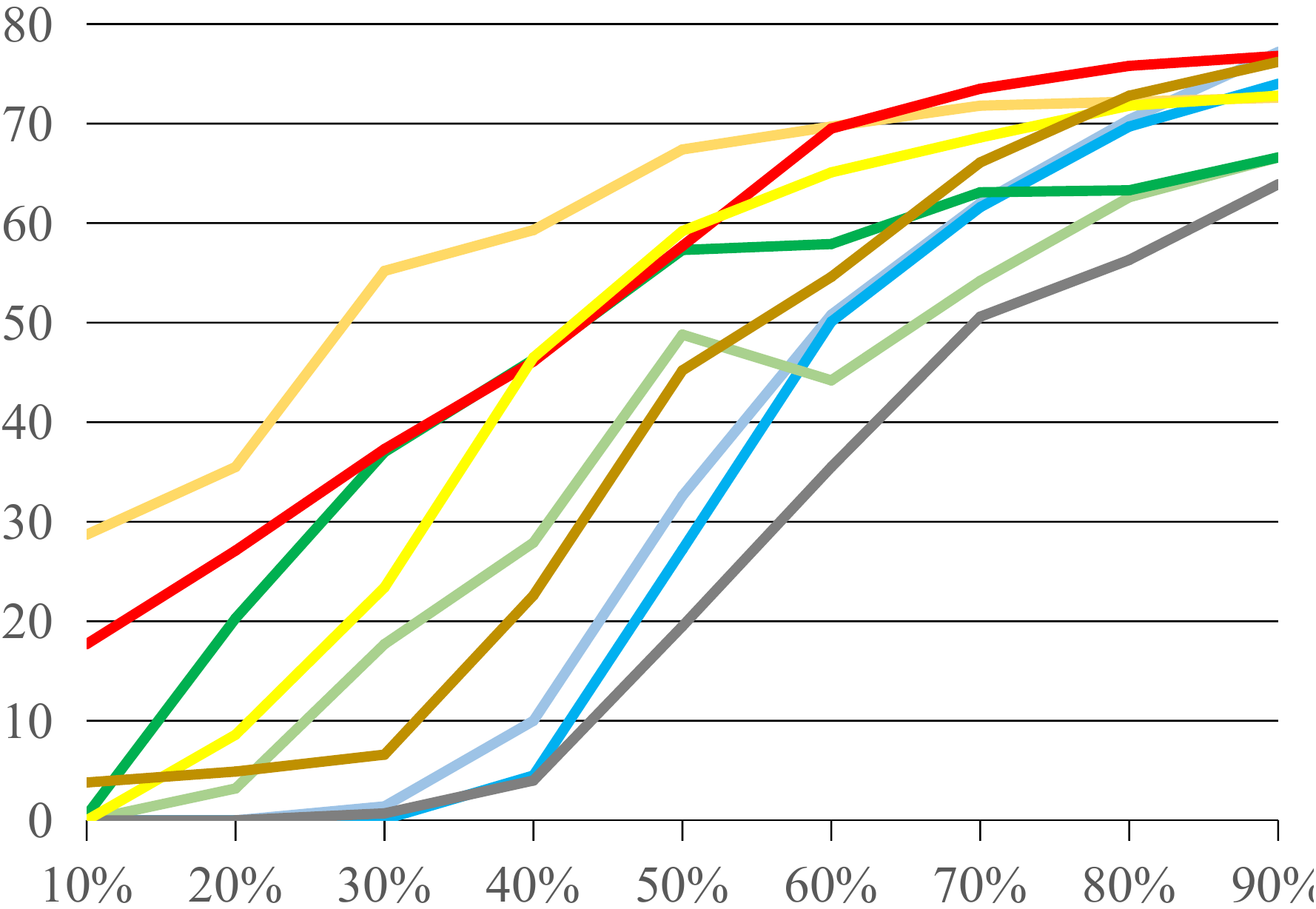}  
    }
  \subcaptionbox*{}{
    \includegraphics[width=0.3\linewidth]{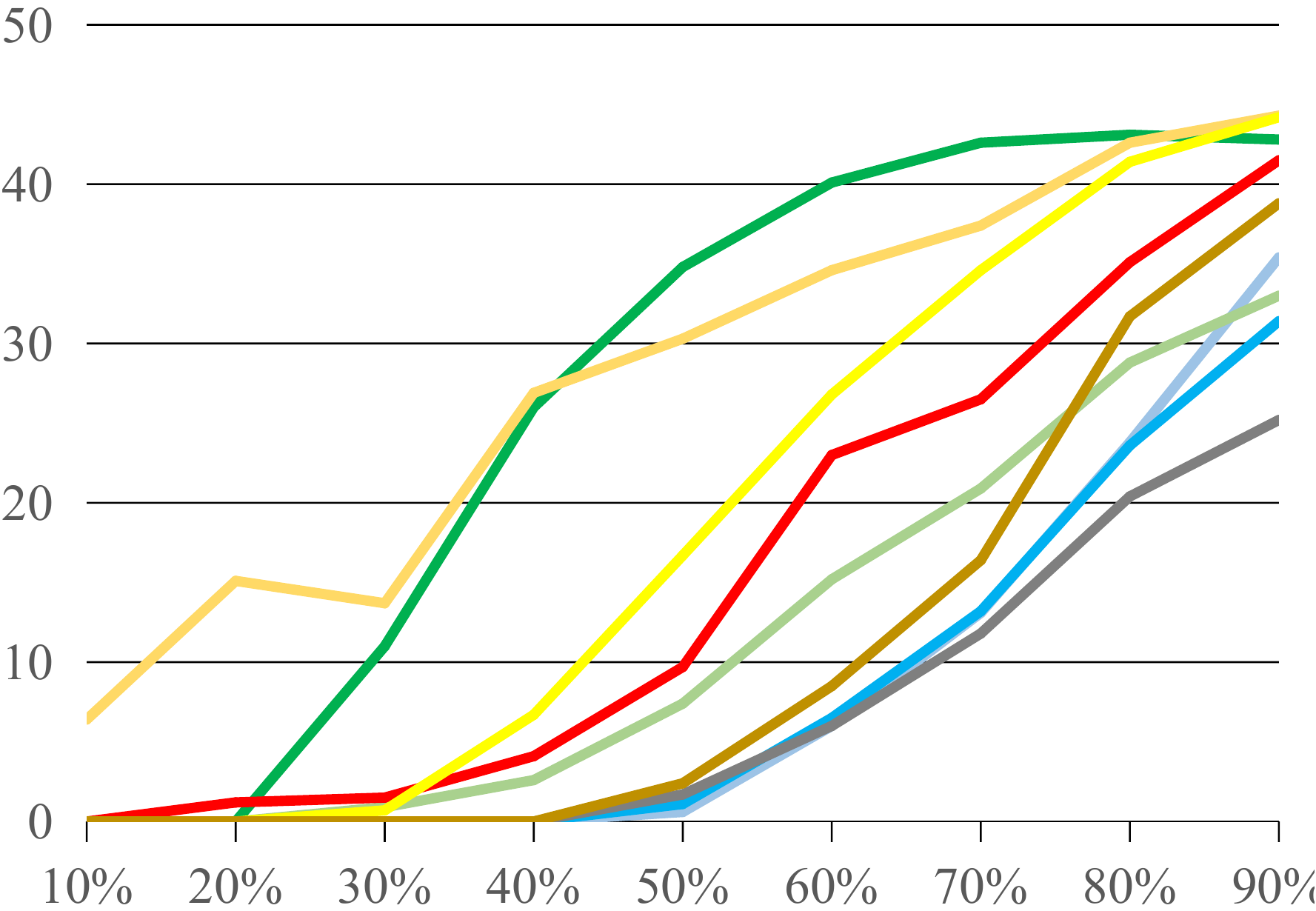}  
    }
  \subcaptionbox*{}{
    \includegraphics[width=0.3\linewidth]{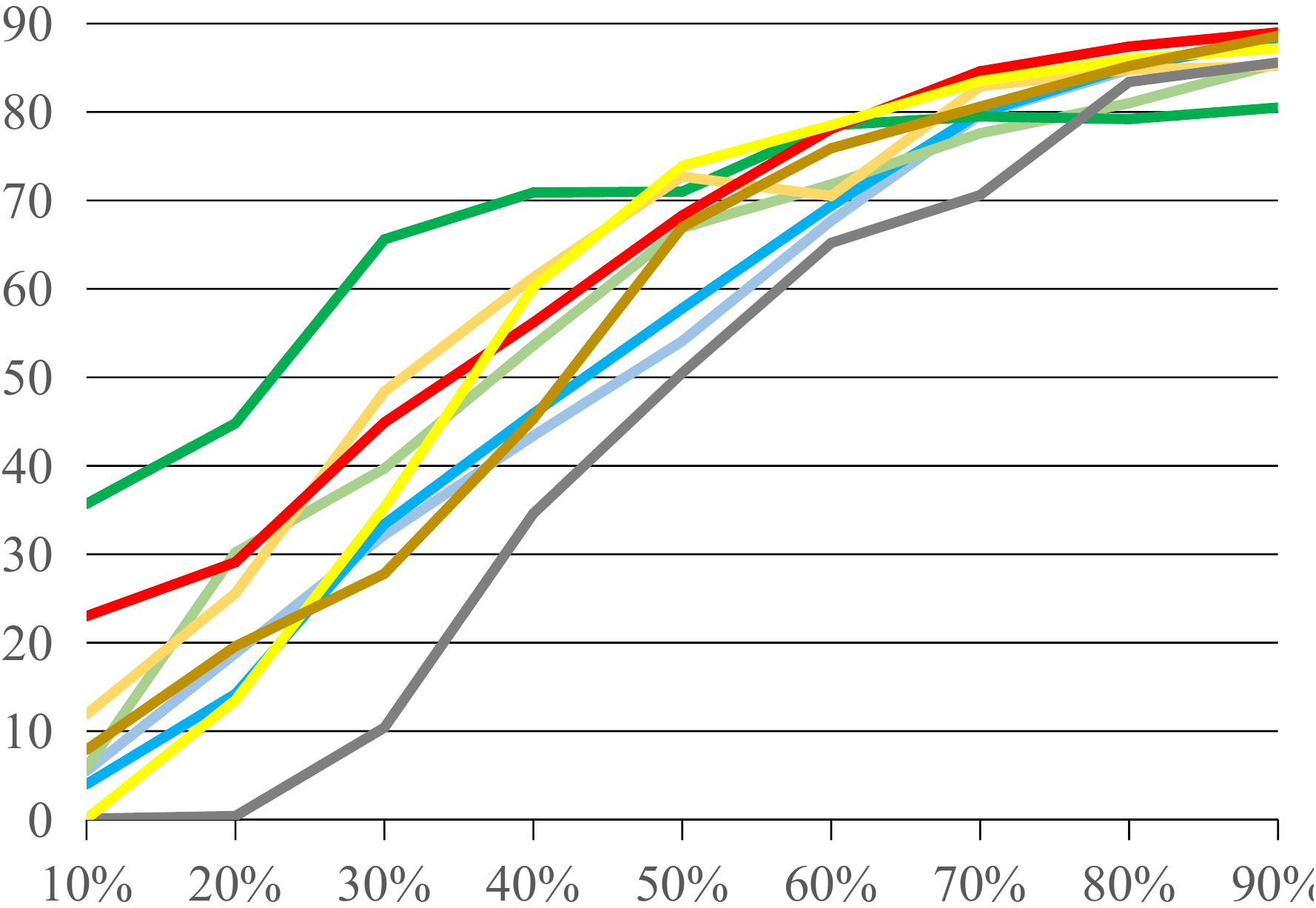}  
    }
  \subcaptionbox*{}{
    \includegraphics[width=0.3\linewidth]{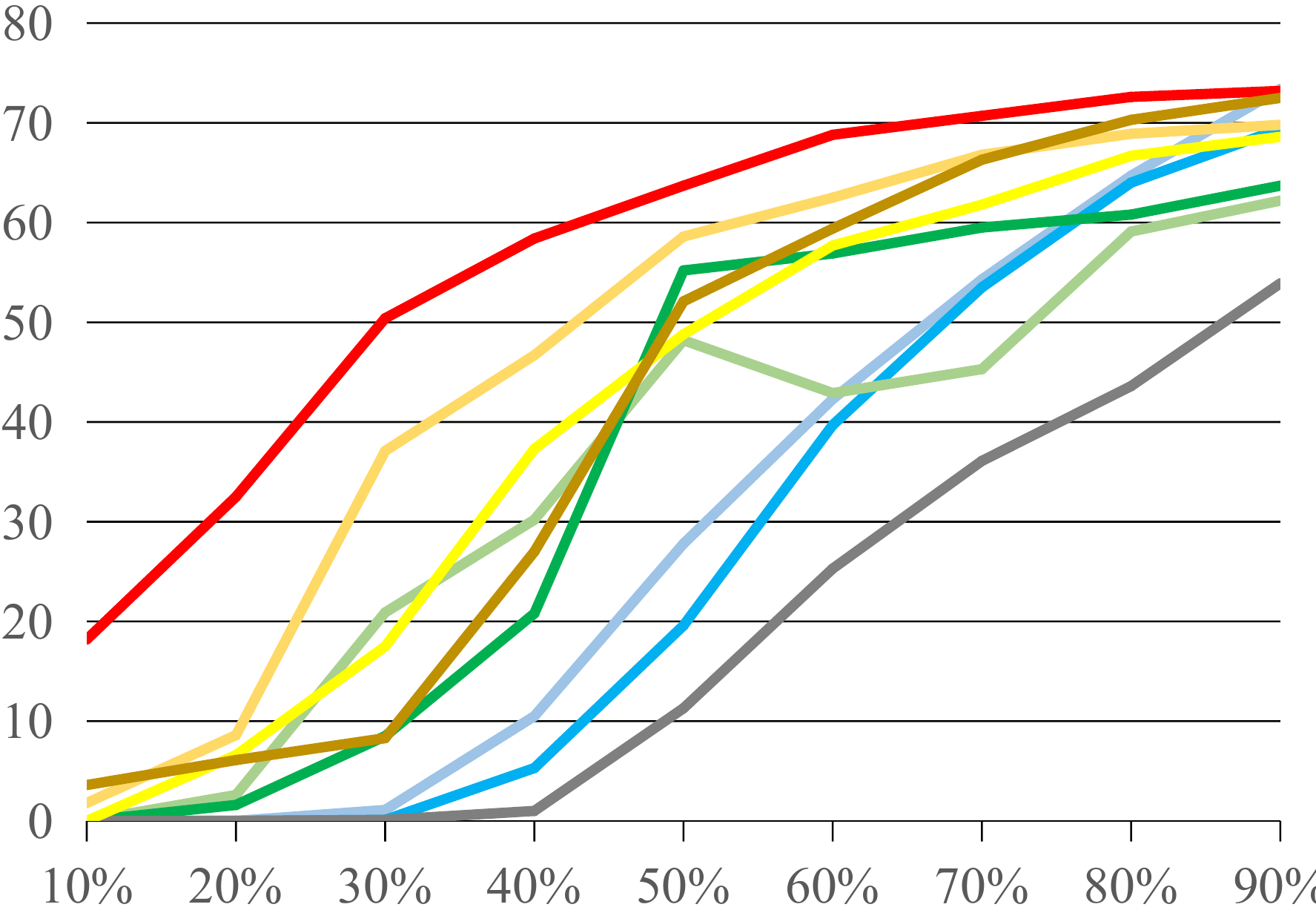}  
    }
  \subcaptionbox*{}{
    \includegraphics[width=0.3\linewidth]{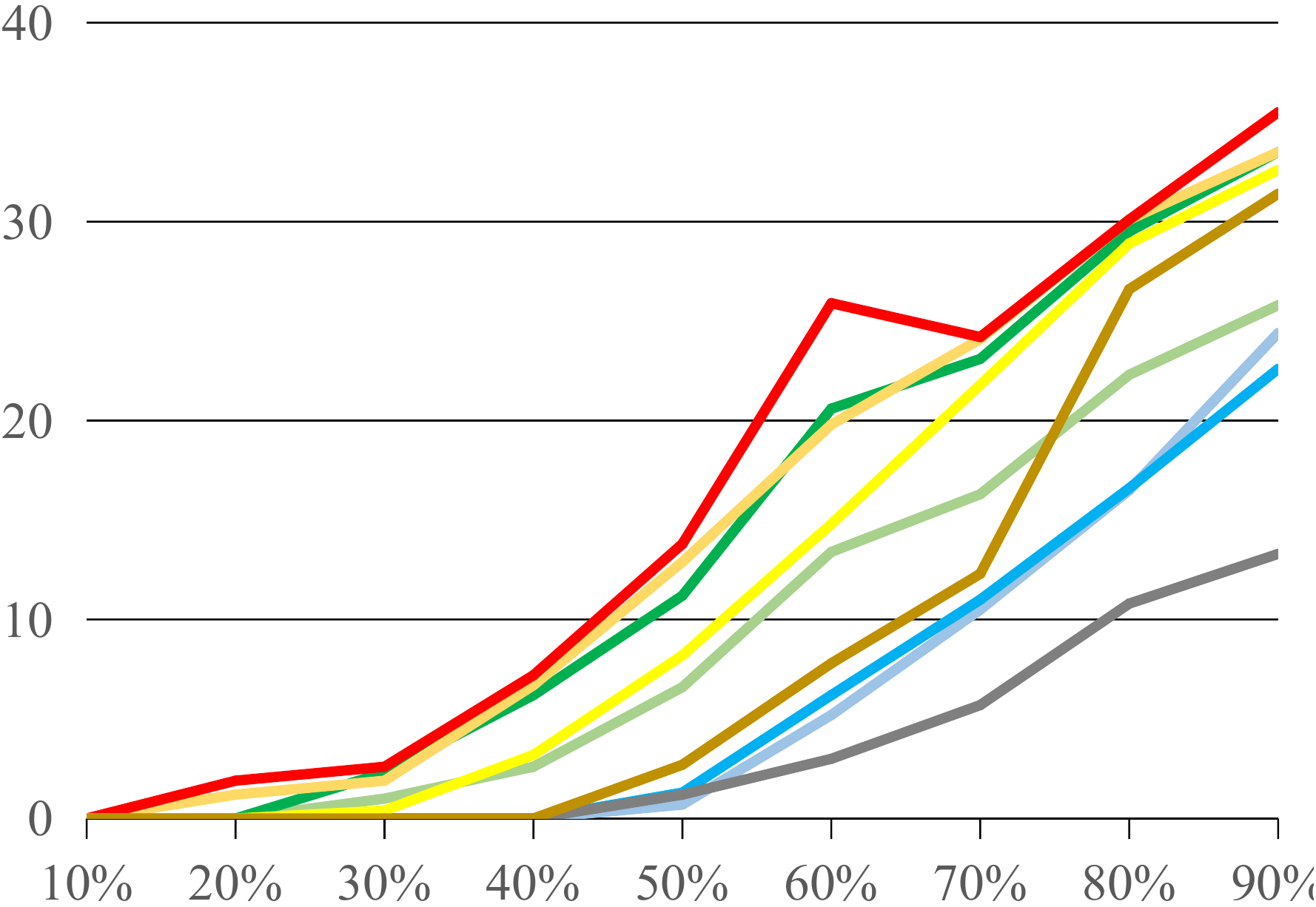}  
    }
  \subcaptionbox*{}{
    \includegraphics[width=0.3\linewidth]{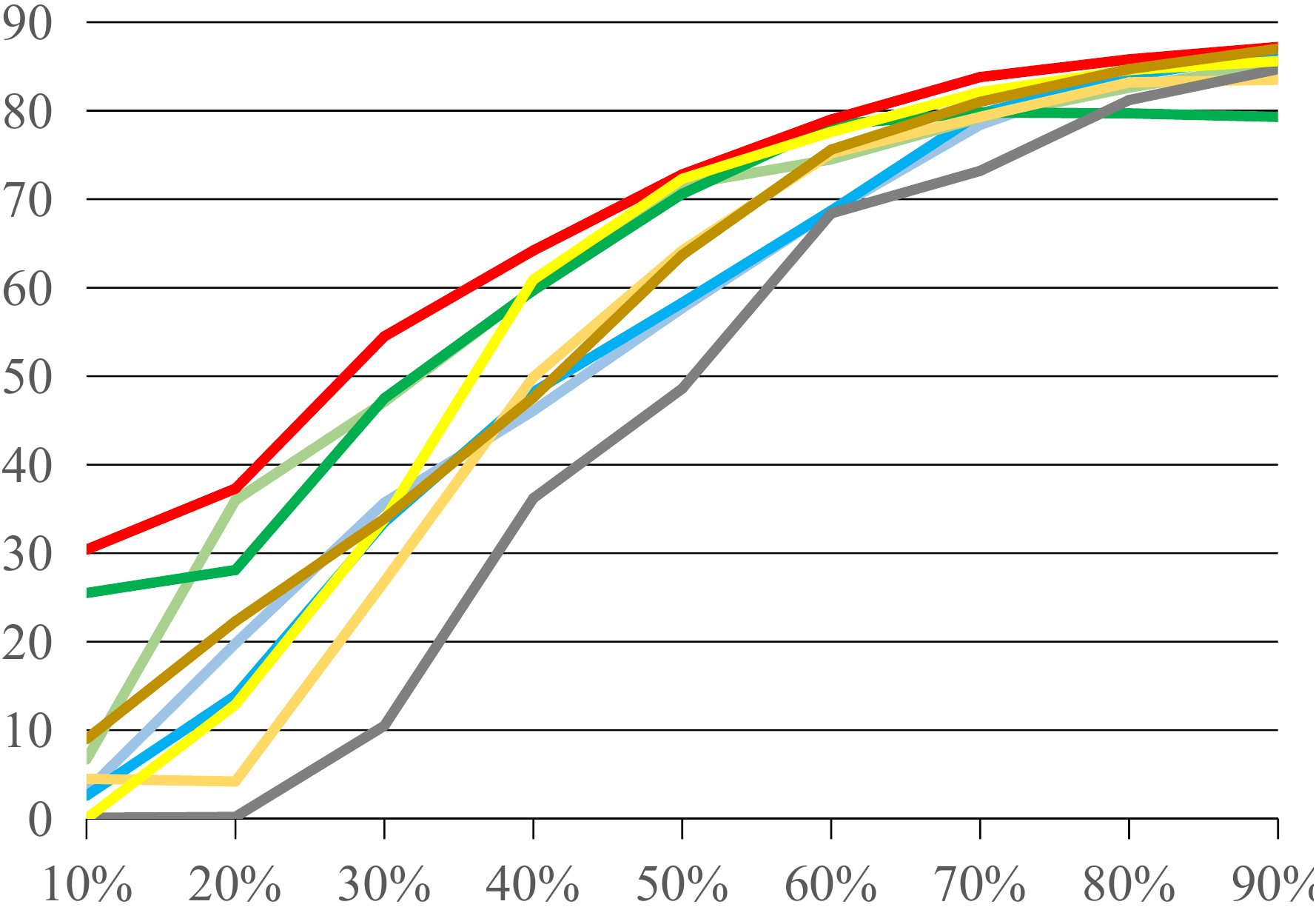}  
    }
  \vspace{-20pt}
  \caption{The OF1 (top) and CF1 (bottom) of our proposed framework and current leading competitors on the settings of partial positive label proportions of 10\% to 90\% on the MS-COCO (left), VG-200 (middle) and Pascal VOC 2007 (right) datasets. Best viewed in color.}
  \label{fig:of1-cf1-result}     
\end{figure*}

\section{Experiments}

\subsection{Experimental Setting}

\noindent{\textbf{Dataset }} Following previous multi-label recognition works \cite{Chen2019SSGRL,Durand2019CVPR,Kim2022LargeLoss}, we conduct extensive experiments on the MS-COCO \cite{Lin2014COCO}, Visual Genome \cite{Krishna2017VG}, and Pascal VOC 2007 \cite{Everingham2010Pascal} datasets for fair comparison. As the most widely used benchmark for multi-label evaluation, MS-COCO contains 120K daily images from 80 object categories where there is 82,801 images as the training set and 40,504 images as the validation set. Considering Visual Genome contains 80,138 categories and most categories have very few samples we select the 200 most frequent categories to obtain a VG-200 subset. Moreover, since there is no train/val split, we randomly select 10,000 images as the test set and the rest 98,249 images are used as the training set. The train/test set will be released for further research. Pascal VOC 2007 contains 9,963 images from 20 object categories, and we follow previous works to use the trainval set for training and the test set for evaluation. 

Due to the fully annotations in these datasets, we randomly drop a certain proportion of positive labels and all negative labels to create the setting of partial positive label learning. To evaluate if our proposed method consistently works in different proportion of partial positive labels, the proportion of known positive labels vary from 10\% to 90\%, resulting in dropped positive labels proportion of 90\% to 10\%.

\noindent{\textbf{Evaluation Metric}} For a fair comparison, we follow previous works \cite{Durand2019CVPR} to adopt the mean average precision (mAP) over all categories for evaluation under different proportions of known labels. And we also compute average mAP over all proportions for a more comprehensive evaluation. Moreover, we follow most previous MLR works \cite{Chen2019SSGRL} to adopt the overall and per-class F1-measure (i.e., OF1 and CF1) for more comprehensive evaluation. Formally, the OF1 and CF1 can be computed by
\begin{gather*}
 OP=\frac{\sum_{i}{N^c_i}}{\sum_{i}{N^p_i}}, CP=\frac{1}{C} \sum_{i}{ \frac{N^c_i}{N^p_i} } \\
 OR=\frac{\sum_{i}{N^c_i}}{\sum_{i}{N^g_i}}, CR=\frac{1}{C} \sum_{i}{ \frac{N^c_i}{N^g_i} } \\
 OF1=\frac{2 \times OP \times OR }{OP+OR},  CF1=\frac{2 \times CP \times CR }{CP+CR}
\end{gather*}
where $N^c_i$ is the number of images that are correctly predicted for the $i$-th label, $N^p_i$ is the number of predicted images for the $i$-th label, $N^g_i$ is the number of ground truth images for the $i$-th label. We also present in detailed the OF1, CF1 results on all known positive label proportion settings.

\noindent{\textbf{Implementation Details }} For fair comparison, we follow previous works \cite{Kim2022LargeLoss, Cole2021ROLE} to adopt the ResNet-101 \cite{He2016ResNet} as the backbone to extract global feature maps $\textbf{f}^n$. We initialize its parameters with those pre-trained on the ImageNet \cite{Deng2009Imagenet} dataset while initializing the parameters of all newly-added layers randomly. We fix the parameters of the previous 91 layers of ResNet-101, and train the other layers in an end-to-end manner. During training, we use the Adam algorithm \cite{Kingma2015Adam} with a batch size of 32, momentums of 0.999 and 0.9, and a weight decay of $5 \times 10^{-4}$. We set the initial learning rate as $10^{-5}$ and divide it by 10 after every 10 epochs. It is trained with 20 epochs in total. For data augmentation, the input image is resized to 512$\times$512, and we randomly choose a number from \{512, 448, 384, 320, 256\} as the width and height to crop patch. Finally, the cropped patch is further resized to 448$\times$448. Besides, random horizontal flipping is also used. $\theta$ is the crucial parameters that control the precision and recall of generating pseudo labels and rejecting noisyl labels. In the training process, the parameters is set to 1 during the first 5 epochs to avoid incurring any noise. Then, they are set to 0.95 at epoch 6 and are decreased by 0.025 for every epoch until it reach 0.6.

\subsection{Comparison with Current Leading Algorithms}
To evaluate the effectiveness of the proposed framework, we compare it with the following algorithms that can be classified into three folds:  (1) \textit{Conventional MLR Algorithms}: \textbf{SSGRL} (ICCV'19) \cite{Chen2019SSGRL}, \textbf{KGGR} (TPAMI'22) \cite{Chen2022KGGR}. Through introducing different visual context structures to guide model learning semantic aware representation for each category, these works achieve state-of-the-art performance on the traditional MLR task. For fair comparisons, we adapt these methods to address the MLR-PPL task by replacing the BCE loss with ``Assume Negative" (AN) loss while keeping other components unchanged. (2) \textit{General Debiased Algorithms}: \textbf{Focal Loss} (ICCV'17) \cite{Lin2017FocalLoss}, \textbf{ASL} (ICCV'21) \cite{Ridnik2021ASL}. These methods focus on operating differently on positive and negative samples to address the positive-negative imbalance problem. Considering the severe imbalance between positive samples and negative samples in the MLR datasets, recent works propose to utilize these algorithms to alleviate this dilemma. (3) \textit{Weakly Supervised MLR Algorithms}: \textbf{CST} (AAAI'22) \cite{Chen2022SST}, \textbf{ROLE} (CVPR'21) \cite{Cole2021ROLE}, \textbf{LL-R} (CVPR'22) \cite{Kim2022LargeLoss}. Specifically, CST transfers knowledge of known labels to generate pseudo labels for unknown labels, ROLE introduces the prior knowledge of average positive label per image to regularize the training process, and LL-R proposes rejecting the large loss samples to prevent model from memorizing the noisy labels. For aforementioned methods, we adopt the same ResNet-101 network as backbone and follow exactly the same train/val split settings. 

\noindent{\textbf{Performance on MS-COCO.}} We first present the performance comparisons on MS-COCO dataset as shown in Table \ref{tab:result}. As the conventional MLR algorithms, SSGRL and KGGR can achieve competitive performance when the positive labels are sufficient, but suffer from severe performance drop when the known positive label proportion decreases to small level. Compared with these methods, the current WSMLR algorithms (i.e., CST, LL-R, ROLE) can maintain great performance when the known positive labels are few, even when the known positive label proportion is 10\%. As presented in Table \ref{tab:result}, our proposed framework obtains the overall best performance over current state-of-the-art algorithms on the settings of 10\%-90\% known positive labels. Specifically, our proposed framework achieves the average mAP of 68.0\%, outperforming the previous best conventional MLR algorithm (i.e., SSGRL) by 9.5\% and the previous best weakly supervised MLR algorithm (i.e., CST) by 7.1\%. It is noteworthy that the proposed framework can achieves more obvious performance improvement when the known positive labels are few. For instance, the mAP improvement over SSGRL and CST algorithms are 29.8\% and 23.9\% respectively when remaining 10\% positive labels. As previously discussed, the OF1 and CF1 are important metrics for evaluating multi-label models, thus we present the OF1 and CF1 results on all known positive label proportion settings in Figure \ref{fig:of1-cf1-result}. As shown, our proposed framework achieves average OF1 and CF1 of 53.5\% and 56.5\%, outperforming the second-best CST algorithm by 14.3\% and 15.9\%.

\noindent{\textbf{Performance on VG-200.}} As previously discussed, VG-200 is a more challenging multi-label dataset because it covers lots of categories and there exists more severe positive-negative imbalance problem in this dataset. Thus, existing methods achieve quite poor performance. As presented in Table \ref{tab:result}, the previous best-performing ASL algorithm merely obtains the average mAP of 32.5\%. In this scenario, by introducing category-adaptive label discovery and noise rejection, our proposed framework still consistently outperforms over current state-of-the-art algorithms on all settings of partial positive labels. Specifically, its average mAP of 37.2\%, outperforming the second-best ASL algorithm by 4.7\%. Compared with current algorithms, we find that our framework obtains the mAP improvement of more than 2.1\% on all known positive label proportion settings.

\noindent{\textbf{Performance on Pascal VOC 2007.}} As the most widely used multi-label dataset, Pascal VOC 2007 often used to evaluate the performance of multi-label model in weakly supervised scenario. As this dataset merely covers 20 categories and contains 5, 000 images as the training set, it is more simple than MS-COCO and VG-200, thus most current algorithms can achieve quite well performance, e.g., the second-best ASL algorithm obtains the average mAP of 86.5\%. However, our proposed framework can still achieve improvement, especially when the known partial labels are few. As shown, our proposed framework improves the average mAP by 0.4\%.

\begin{table}[!h]
  \centering
  \begin{tabular}{c|ccc}
  \hline
  \centering \diagbox{Methods}{Datasets} & MS-COCO & VG-200 & VOC2007 \\
  \hline
  \hline
  \centering Ours CSSL & 62.8 & 32.3 & 87.4 \\
  \centering Ours CALD & 69.8 & 35.7 & 88.2 \\
  \centering Ours CALD w/o CATU & 68.1 & 34.0 & 87.5 \\
  \centering Ours CANR & 72.4 & 37.3 & 88.2 \\
  \centering Ours CANR w/o CATU & 71.7 & 36.1 & 87.9 \\
\hline
  \hline
  \centering Ours & \textbf{74.1} & \textbf{39.1} & \textbf{88.6} \\
  \hline
  \end{tabular}
  \caption{Comparison of mAP of our framework merely using CSSL module (Our CSSL), our framework merely using CALD module (Our CALD), our framework merely using CALD module without CATU (Ours CSLC w/o CATU), our framework merely using CANR module (Ours CANR), our framework merely using CANR module without CATU (Ours CANR w/o CATU) and our framework (Ours) on the MS-COCO, VG-200 and Pascal VOC 2007 datasets when the partial positive label proportion is 50\%.}
  \label{tab:ablation-result}
\end{table}

\subsection{Ablative Studies}
As previously discussed, we introduce a category-specific semantic learning (CSSL) module that guides model learning semantic representation of each category by incorporating category semantics. To verify the contribution of this module, we conduct experiments that merely use this module (namely, Ours CSSL) and compare it with the baseline algorithm ResNet-101 on the MS-COCO, VG-200, and Pascal VOC 2007 datasets. As presented in Table \ref{tab:ablation-result}, it obtains mAP of 62.8\%, 32.3\%, and 87.4\% on these three datasets when the partial positive label proportion is 50\%, with the mAP improvements of 16.8\%, 6.1\%, and 6.8\%.

Considering recent works mainly propose different kinds of algorithms to learn category-specific visual representation, we treat CSSL as the baseline method and compare it with our proposed framework to verify the contribution of category-adaptive label discovery and noise rejection as a whole. As shown, by introducing category-adaptive label discovery and noise rejection modules, our proposed framework improves the mAP to 74.1\%, 39.1\%, and 88.6\%, with an improvement of 11.3\%, 6.8\%, and 1.2\% respectively.

Since the proposed framework consists of two complementary modules, i.e., category-adaptive label discovery (CALD) and category-adaptive noise rejection (CANR) modules, in the following we will conduct more ablative experiments to analyze the separate contributions of these two modules in detail.

\noindent{\textbf{Analysis of the CALD module}} 

To analyze the actual contribution of the CALD module, we conduct experiments that merely use this module (namely, Ours CALD) and compare it with the baseline algorithm CSSL on the MS-COCO, VG-200, and Pascal VOC 2007 datasets. As shown in Table \ref{tab:ablation-result}, it obtains mAP of 69.8\%, 35.7\%, and 88.2\% on the MS-COCO, VG-200, and Pascal VOC 2007 datasets when the partial positive label proportion is 50\%, with the mAP improvement of 7.0\%, 3.4\%, and 0.8\%, respectively. 

\noindent{\textbf{Analysis of the CANR module}} 

To analyze the actual contribution of the CANR module, we conduct experiments that merely use this module (namely, Ours CANR) and compare it with the baseline algorithm CSSL. As presented in Table \ref{tab:ablation-result}, it obtains mAP of 72.4\%, 37.3\%, and 88.2\% on the MS-COCO, VG-200, and Pascal VOC 2007 datasets, with the mAP improvement of 9.6\%, 5.0\%, and 0.8\%, respectively.

\noindent{\textbf{Analysis of Category-Adaptive Threshold Updating}}  
In the CALD module, the crucial parameter $\theta^{pos}_c$ controls the precise and recall of generating pseudo labels. Considering it is impractical and exhausting to find a best value for different datasets and different settings, we propose category-adaptive threshold updating to adaptively adjust $\theta^{pos}_c$ for each category $c$. To verify its contribution, we conduct an experiment to compare with the baseline using a manually tuned $\theta^{pos}_c$. As presented in Table \ref{tab:ablation-result}, it decreases mAP from 69.8\%, 35.7\%, and 88.2\% to 68.1\%, 34.0\%, and 87.5\%, respectively.

Similar with the CALD module, the CANR module contains two crucial parameter controls the precise of rejecting noisy labels, i.e., $\theta^{pos}_c$ and $\theta^{neg}_c$. As we adopt category-adaptive threshold updating algorithm to update these parameters, we conduct an experiment to compare with the baseline using a manually tuned $\theta^{pos}_c$ and $\theta^{neg}_c$ to verify its contribution. As shown, removing the category-adaptive threshold updating decreases mAP from 72.4\%, 37.3\%, and 88.2\% to 71.7\%, 36.1\%, and 87.9\%, respectively.

\section{Conclusion}
In this work, we propose a novel perspective to train multi-label model with partial positive labels by exploring category-specific cross-image semantic correlation. To achieve this, we design a  unified framework, Category-Adaptive Labels Discovery and Noise Rejection, that consists of two complementary modules (i.e., CALD and CANR modules) to complement unknown labels and discards noisy labels. Besides, we propose a novel category-adaptive threshold updating that adaptively adjusts the threshold of each category that avoids extremely time-consuming and laborious manual tuning in different datasets and different known positive label proportions. We conduct extensive experiments on several large-scale MLR datasets, including Microsoft COCO, Visual Genome and Pascal VOC 2007, to demonstrate the effectiveness of our proposed framework.

\bibliographystyle{IEEEtran}
\bibliography{reference}

\end{document}